\documentclass[conference]{IEEEtran}
\IEEEoverridecommandlockouts
\usepackage{cite}
\usepackage{amsmath,amssymb,amsfonts}
\usepackage{algorithmic}
\usepackage{graphicx}
\usepackage{textcomp}
\usepackage{xcolor}
\def\BibTeX{{\rm B\kern-.05em{\sc i\kern-.025em b}\kern-.08em
    T\kern-.1667em\lower.7ex\hbox{E}\kern-.125emX}}
\begin{document}

\title{TinyMLOps: Operational Challenges for Widespread Edge AI Adoption}

\author{
    \IEEEauthorblockN{Sam Leroux\IEEEauthorrefmark{1}\IEEEauthorrefmark{2}, Pieter Simoens\IEEEauthorrefmark{2}, Meelis Lootus\IEEEauthorrefmark{1}, Kartik Thakore\IEEEauthorrefmark{1}, Akshay Sharma\IEEEauthorrefmark{1}}
    \IEEEauthorblockA{\IEEEauthorrefmark{1}hotg.ai
    \\\{sam, meelis, kartik, akshay\}@hotg.ai}
    \IEEEauthorblockA{\IEEEauthorrefmark{2}Ghent University - imec
    \\\{sam.leroux, pieter.simoens\}@ugent.be}
}

\maketitle

\begin{abstract}
Deploying machine learning applications on edge devices can bring clear benefits such as improved reliability, latency and privacy but it also introduces its own set of challenges. Most works focus on the limited computational resources of edge platforms but this is not the only bottleneck standing in the way of widespread adoption. In this paper we list several other challenges that a TinyML practitioner might need to consider when operationalizing an application on edge devices. We focus on tasks such as monitoring and managing the application, common functionality for a MLOps platform, and show how they are complicated by the distributed nature of edge deployment. We also discuss issues that are unique to edge applications such as protecting a model's intellectual property and verifying its integrity.
\end{abstract}


\section{Introduction}
TinyML \cite{warden2019tinyml}, referring to the use of machine learning (ML) models on resource constrained edge devices, is an attractive deployment paradigm for various applications such as smart home appliances, virtual assistants, autonomous vehicles and smart surveillance. For these applications, edge deployment can provide a lower latency, increased robustness, scalability and privacy compared to deployment scenarios where the model is evaluated on cloud infrastructure \cite{dutta2021tinyml}. There are however several challenges that hinder large scale adoption of edge AI. The most obvious obstacle is the limited processing power of edge devices. As these devices are typically wearable and battery powered, intrinsic constraints on energy consumption, form factor and heat dissipation limit their computational resources. Most of the work in the field of TinyML focuses on improving the efficiency of the models on resource constrained devices. 
In this paper, we focus less on this computational aspect, but instead investigate what other challenges arise when edge ML is put in production and what is being done to solve them. We do not concentrate on ML model design or training but rather on the operational side. We also do not limit ourselves to a specific application (e.g. image recognition, natural language processing, ...) as the discussed approaches are applicable to all these use cases. We implicitly assume that the ML model is a deep neural network as these types of models are currently the state of the art for various machine learning tasks but this is not a limitation for most of the discussed approaches, as they can be useful for other families of ML models as well.
\\
\newline
The remainder of the paper is structured as follows.  We begin in section \ref{sec:processing} with a discussion on the computational aspect of TinyML. In section \ref{sec:tinymlops}, we then discuss MLOps and introduce the idea of TinyMLOps, a set of additional challenges and considerations that pop up when an ML model is deployed on edge devices instead of in the cloud for production. In the subsequent sections, we address very specific challenges such as dealing with a fragmented device landscape (section \ref{sec:target}), protecting the intellectual property of an ML model (section \ref{sec:ip}) and validating the result of a model (section \ref{sec:verify}). We conclude in  section \ref{sec:conclude} and give some pointers for future work.

\section{The computational aspect of TinyML}
\label{sec:processing}
No paper discussing the challenges of TinyML would be complete without an overview of techniques that can be used to improve the efficiency of machine learning models on resource constrained devices. Despite the short history of TinyML, there is already a vast amount of literature on approaches that reduce the computational cost, memory footprint and energy consumption of ML models targeting edge deployment. These techniques include pruning \cite{han2015learning}, model quantization \cite{nagel2021white}, knowledge distillation \cite{yang2019snapshot}, adaptive computation \cite{leroux2018iamnn} or automated neural architecture search \cite{zoph2016neural}. Several of these techniques can only reach their full potential if they have the support of the hardware platform they are deployed on \cite{sze2017efficient}. There is therefore a vast opportunity for hardware-software co-design to achieve the highest possible efficiency \cite{moons201714}. As the computational aspect of TinyML is not the focus of this paper, we instead refer to some excellent survey works on this topic \cite{sze2017efficient, xu2020edge, verhelst2017embedded, dutta2021tinyml}.

\section{MLOps and the need for TinyMLOps}
\begin{figure*}
    \centering
    \includegraphics[scale=0.35]{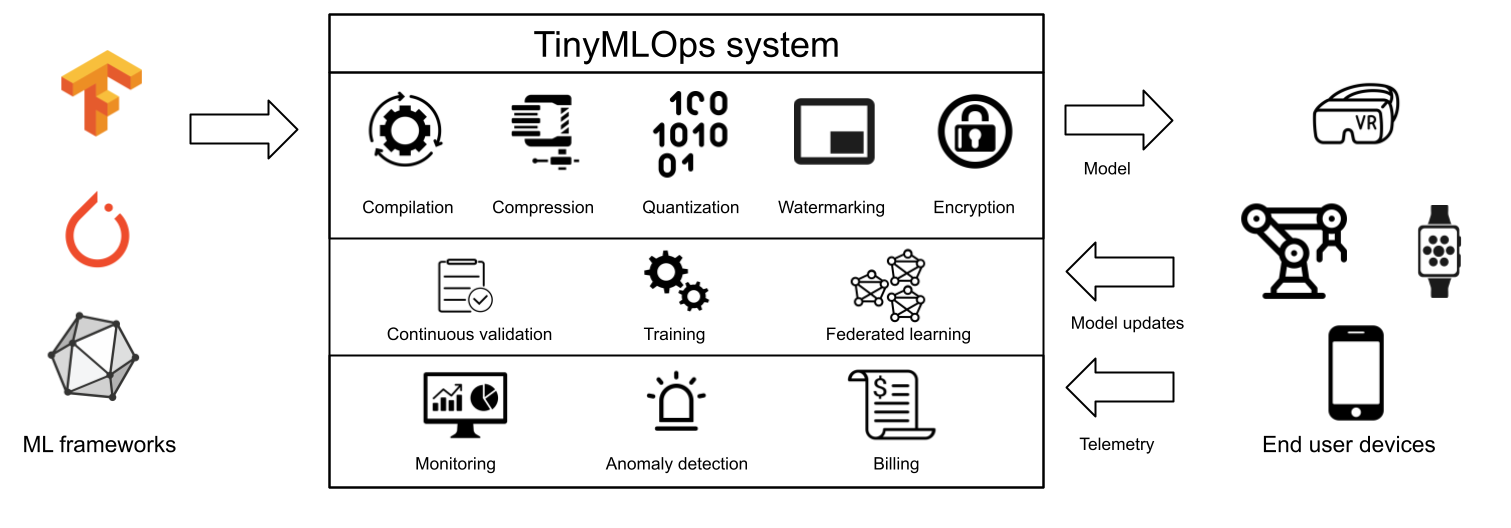}
    \caption{Overview of the different functionality of a TinyMLOps system}
    \label{fig:my_label}
\end{figure*}

\label{sec:tinymlops}
In the past decade, we have seen an explosion of machine learning techniques being applied in various domains such as finance, medicine, commerce or entertainment. These models are typically developed by specialized data scientists, usually in a relatively ad hoc manner. The result is a complicated pipeline of multiple scripts and frameworks that makes it very difficult to deploy the model in production. An often quoted statistic is that up to 85\% of corporate AI initiatives are struggling to move beyond the test stages \cite{gartner}. While this number is only an estimation, it is fair to say that a successful integration of machine learning models in a complex user-facing application requires a lot of forethought and continuous maintenance, especially if the model is updated over time with new training data. In 2015, authors from Google published a paper discussing the technical debts of machine learning systems \cite{sculley2015hidden}. ``Technical debt'' is a term from software engineering that refers to the additional future cost of maintaining an application caused by cutting corners during development. Some typical causes include lack of documentation, duplicated code or tight coupling of components \cite{suryanarayana2014refactoring}. Adding machine learning functionality to the application only increases the risk as the system can now be updated in many ways, in addition to updating the source code. New training data, better feature definitions and hyper parameter settings or new model architectures all trigger an update of the application. It is often difficult to keep track of what dependencies a model has, what the performance of the model is or to detect when it goes wrong and the model is making (potentially expensive) mistakes \cite{muralidhar2021using}. All of these concerns gave rise to the field of MLOps. Being an extension of DevOps, MLOps brings software development practices such as automation, monitoring, integration and testing to ML applications. As the field of machine learning is maturing and companies see a clear return on investment from their ML-based products, MLOps is becoming more and more widespread. A recent report even estimated the MLOps market to reach an annual revenue of \$4 billion by 2025 \cite{deloite}.
\\
\newline
In this paper, we argue for the extension of MLOps to TinyMLOps. As we move from a centralized, cloud based application to a decentralized edge based deployment, new challenges will arise that need to be dealt with to support successful adoption of TinyML. Figure \ref{fig:my_label} illustrates some of the functionality of such a TinyMLOps system. We will discuss these challenges in the next sections.

\subsection{Managing model versions}
In a centralized, cloud based application, it is probably sufficient to have a single model for all users, we simply use the latest and greatest architecture for the task at hand. In a decentralized, edge based application, the model will be deployed to the end user's device. Different users will have different devices with different computational resources, storage availability and network connectivity. Instead of training a single model, we might need to support multiple models, each with their own computational cost and accuracy trade off \cite{leroux2019multi}. This allows us to push a smaller, more efficient model to a device with limited resources or a large, more accurate model to a more powerful device. To make this even more complicated, the best model for a given device might depend on external factors besides the computational resources of the device. If the device is connected to an external power supply, energy consumption might be less of an issue compared to when it is unplugged and has to rely on battery power. This might mean that a different model could be preferred, depending on the battery level. Similarly, depending on the situation, the user might prefer a slower, more accurate model or a faster, less accurate model or even a model that is fast to download on a slow network connection compared to a larger model when he is connected to WiFi. 
\\
\newline
A common technique for TinyML model optimization is to use reduced precision operations. Traditionally, neural networks relied on 32 bit numerical operations for training and evaluation. It was found however that inference can work fine with 8 bit \cite{vanhoucke2011improving}, 3 bit \cite{nayak2019bit}, 2 bit \cite{venkatesh2017accelerating}  or even 1 bit (binary) weights and operations \cite{courbariaux2016binarized}. Training a model is more difficult but models have been successfully trained using 4 bit weights \cite{nagel2021white}. Low precision operations however do not necessarily guarantee faster and more efficient models on all hardware platforms. Special support from hardware is needed to obtain an increased throughput or reduced energy consumption \cite{judd2016stripes}. As different hardware platforms might support a different set of operations and bit widths, we might need to develop different versions of a model, each targeting a certain platform. 
\\
\newline
All of these examples show that the number of models that need to be managed by a TinyMLOps system is much larger than the number of models for a corresponding centralized deployment. Existing solutions for storing models in a centralized repository will therefore have to be extended to track the relationship between different versions of the models, recording what optimizations are applied to every instance. If the base model is updated or retrained, we also have to automatically trigger the execution of the optimization pipeline that generates different quantized or pruned versions of the base model. 
\\
\newline
In addition to the model itself, the machine learning pipeline will also require data preprocessing and postprocessing operations such as normalization, thresholding or even some control logic to activate a different part of the pipeline depending on the result of a first model. The TinyMLOps system should make it easy for users to configure pipelines like this. A promising approach is using WebAssembly\cite{haas2017bringing} to package these different operations in portable and re-usable modules \cite{lootus2020vm}.
\subsection{Observability}
\label{sec:observability}
As ML models are increasingly being used in large consumer-facing applications, monitoring and observability are key to ensuring that a model keeps performing as expected \cite{shankar2021towards}. There are several ML observability solutions available \cite{arize, whylabs, evidently} that typically monitor the distribution of input values to detect data drift. This allows machine learning engineers to detect model performance degradation early on. It is relatively straight-forward to implement model observability when the model is deployed on a cloud platform as all the input data is sent to a centralized location. Here, we can easily track the input data distribution or perform anomaly detection. We can also store anomalous data points for analysis or retraining the model. It becomes less trivial if the model is deployed on the edge. A major benefit of edge deployment is that no data needs to leave the device, which provides stronger privacy guarantees than cloud processing. But this argument would be rendered invalid if we periodically shared data with the cloud for purposes of analysis and retraining. We could record some basic statistics on the data locally and share these with the cloud in an anonymized way but it depends on the application if this is useful or not. In addition, we are also interested in monitoring the number of requests a user has made and the execution time of the model. As different users will have different hardware platforms, offering different computational resources, it is very interesting to record the actual execution time, memory and energy consumption on the end-user's device. These types of telemetry data can signal performance issues that might necessitate further performance optimizations of the model.  We might decide to store these statistics locally and transmit them to the cloud when the device is connected to WiFi or we can ask the user's permission to share these. All of these considerations have to be taken into account when implementing a TinyMLOps observability functionality. 

\subsection{Pay-per-query business model}
A common business model for ML applications is pay-per-query. Google's Cloud vision API for example charges users \$1.50 per 1.000 requests for tasks such as face detection \cite{cloudvision}. Similar to observability, it is trivial to implement this in a cloud API as all requests have to be processed by the same endpoint in an online fashion. If we decide to deploy the model to the end-user device however, this becomes more difficult. While it is true that the end user now carries the computational cost of model inference, the application developer still needs to be compensated for the cost of developing and training the model. A pay-per-query business model might still be a viable business model in the case of distributed edge deployment. This however is much more difficult to implement as the model is now replicated on a large number of end-user's devices that might not even be connected to the internet the moment they are evaluating the model. We could offer prepaid packages where the user purchases the right to perform a certain number of model calls. The application then needs to keep track of how many requests the user has executed and will deny access if this exceeds the number of requests the user has paid for. Doing this in a secure offline way on untrusted hardware is however not trivial \cite{christodorescu2020towards} and would be a very useful feature for a TinyMLOps solution.

\subsection{Retraining and personalizing models}
Modern machine learning applications are not static anymore, they are updated continuously as new data has been observed. This allows them to deal with shifts in the input data and benefits the overall performance of the system. There are some challenges such as dealing with catastrophic forgetting \cite{thai2021does} when designing machine learning models that support continuous learning but a centralized, cloud based solution makes it much easier to do this effectively than a distributed edge based approach. In a cloud scenario, all data is transmitted to the cloud for processing. Although labor intensive, it is relatively easy to store this data, clean it and potentially annotate it for training new iterations of the model. Once an updated model is trained, we can validate it on a hold-out dataset and replace the existing model if its accuracy is satisfactory. In an edge based application, this is far from trivial to do. A major benefit of an edge based application is that all data stays on the local device. This is much more attractive from a privacy point of view but this also means that it is impossible to centralize all the data to train new models without invalidating this privacy argument. Instead, we will have to rely on Federated Learning to update the model \cite{mcmahan2017communication}. With Federated Learning, a user downloads the current model and updates it locally with his own data. The updates to the model are then sent to the cloud where it is averaged with other user's updates to improve the shared model. Periodically, the updated global model is pushed to each device, allowing individual users to benefit from the experience of the model on other devices. Federated Learning is not a trivial problem as the heterogeneous (non-iid) data generated by different clients causes the local models to diverge, making it difficult to aggregate the local updates into a global update \cite{zhu2021federated}. Despite being a relatively new research field, Federated Learning is already deployed in practice \cite{bonawitz2019towards}. 
\\
\newline
There are several additional challenges that arise when implementing Federated Learning in a TinyML setting. The first one is the computational cost. As we now rely on the edge device to calculate model updates instead of just performing inference using a trained model, the computational cost increases accordingly \cite{luo2021cost}. In addition, model updates need to be shared with the cloud backend periodically. This will have a direct impact on the energy consumption of the application. It might be possible to temporarily store some of the data locally and to calculate the model updates when the device is idle or connected to a charger. Several techniques have been developed to reduce the communication overhead of the Federated Learning techniques \cite{mcmahan2017communication, yao2019federated, chen2019communication, yao2018two, mills2019communication, xu2020ternary}. This is especially useful when Federated Learning is used in wireless sensor nodes as network communication is expensive in terms of energy consumption. Other techniques have reduced the computational cost for the individual nodes \cite{duan2019astraea, lim2020federated, semwal2020fedperf}. 
\\
\newline
Most Federated Learning approaches make the assumption that labelled data is available \cite{diao2021semifl}. This might be realistic in certain settings such as a medical use case where individual doctors annotate medical records but this is not very realistic for a TinyML setting. Here, the individual nodes might operate without human intervention or feedback which means that the data remains completely unlabeled. Even if the user can be asked to provide a label, most likely, this will be of low quality \cite{jin2020towards}. Several techniques have been developed that can use unlabelled local data to improve the global model either in a semi-supervised \cite{diao2021semifl, jeong2020federated, albaseer2020exploiting, itahara2020distillation} or unsupervised way \cite{zhang2020federated, tzinis2021separate} .
\\
\newline
We do not necessarily need to aggregate all the local updates to the model into a global model. By deploying a model to an edge device, the model will usually only process data belonging to a single user or generated in a single location. We could exploit this to train specialized models that are ``overfitted'' to a specific user or location. An example of this would be a personalized auto complete functionality \cite{kannadasan2019personalized} or an anomaly detection model trained for predictive maintenance that over time learns the characteristics of a single machine or sensor \cite{anaissi2021personalized, LEROUX2022107763}. 
 
\section{Targeting a fragmented IoT landscape}
\label{sec:target}
The cloud infrastructure used for ML tasks is relatively uniform. Usually, these are high-end Linux servers equipped with NVIDIA hardware accelerators. The edge landscape is however much more fragmented with a wide range of different devices from different vendors, each with different software support and hardware capabilities. As we are seeing a trend towards custom hardware accelerators that support ML applications in a more energy efficient way \cite{li2020survey}, we expect that this fragmentation will only increase in the future.  This clearly is a major challenge for a TinyMLOps platform. To deploy the application on a new device, we will first need to check that all required operations are supported by the underlying platform. We then might need to quantize the model or perform further vendor-specific optimizations. At the same time, ML engineers might be using different software frameworks to develop their models. Not every framework will support every deployment target and a lack of standardization in this area will make it very difficult for developers to support the increasing number of different types of devices. There are some efforts such as the Open Neural Network Exchange format (ONNX)\cite{onnx} and the Neural Network Exchange Format (NNEF)\cite{nnef} that represent and interchange neural networks among deep learning frameworks and inference engines. Different development frameworks can export trained models to these formats that can then be loaded and executed on different hardware platforms. Since these formats are relatively new and the ML field moves quickly, not all operations are readily supported by these formats and it is not trivial to use them for more exotic or novel models. Apache TVM \cite{chen2018tvm} is similar in the sense that it supports models developed in different frameworks. These are then compiled into minimum deployable modules for multiple hardware platforms. TVM can perform different optimizations such as operator fusion and data layout transformation to improve the performance on a certain hardware platform. Other related frameworks are Intel's OpenVino \cite{gorbachev2019openvino} or Antmicro's Kenning \cite{kenning}.
\\
\newline
Modern cloud applications are developed using a cloud native approach emphasizing agility. Software is packaged in loosely-coupled microservices running as containers. This allows developers to easily update and scale their application. These principles can also be applied to edge computing where virtualization can support portable machine learning containers \cite{lootus2020vm}. These containers could then easily be deployed to different target devices, solving the fragmentation issue. By running the containers in an isolated sandbox, we can restrict the access to parts of the operating system or external sensors, improving the security of the whole system. As a next step, the containers could be controlled by an orchestration framework that automatically deploys updated models or that distributes an application over multiple devices \cite{de2018dianne}. We could then envision a marketplace where every device in the network can potentially execute a certain machine learning workload. Depending on the requirements, a certain target is chosen and the container is transmitted to that device for execution. Owners of the device will be incentivized to run workloads as they receive a monetary compensation. A smartphone app for example could decide to offload its computations to the powerful GPU of a self-driving car while the user is inside. Or it could seamlessly move the container to an edge server if it is available \cite{zhu2020toward}. This virtualization could also enable hybrid edge-cloud applications where, depending on the available resources, the model is evaluated on edge or cloud hardware. It is even possible to split a model between edge and cloud \cite{leroux2017cascading, banitalebi2021auto, grulich2018collaborative, tuli2021splitplace}.

\section{Protecting the model's intellectual property}
\label{sec:ip}
A trained machine learning model can represent a significant intellectual value for the owner. Part of this value comes from the highly specialized skill set of the developer who has spent a large amount of trial and error developing and tuning the model, especially if the model needs to be optimized for edge deployment. In addition, training the model itself requires access to powerful compute resources. It is not uncommon to train models for days or weeks on high-end GPU hardware. As an extreme example, training OpenAI's GPT-3 language model on cloud infrastructure is estimated to cost \$4.6M \cite{gpt3}. Finally, as the key to better performing models is usually a large, high quality training data set, developers often spend large amounts of time collecting, cleaning and manually labeling data. All of this adds up to a significant investment that is needed before a model is ready for production. It is therefore not hard to imagine that unscrupulous actors might try to steal the trained model to use it for their own purposes without limitations, circumventing a pay-per-request business model or even to set up a plagiarised service, offering payed access to the model.  
\\
\newline
For the purpose of this paper, we distinguish between two threat models: direct and indirect ML model stealing. With direct model stealing, the attacker is somehow able to obtain the exact trained weights of the model. He can then use these to instantiate his own copy, circumventing any rate restrictions or pay-per-request billing. He might even be able to extract training data that could be privacy sensitive (e.g. in a medical application) \cite{shokri2017membership, fredrikson2015model}. Even if the attacker is unable to obtain the exact weights of the model he might still be able to extract valuable intellectual property using indirect model stealing. Here, the trained model is considered a black box and the attacker can only query the model. But by making repeated queries to the model, each time providing an input data point and recording the prediction of the model, he is able to construct a labelled data set over time. He can then use this data to train a machine learning model of his own that mimics the behaviour of the original model. While the obtained weights or architecture might differ from the original model, this student-teacher learning \cite{yang2019snapshot} approach can allow the attacker to train a similar model for a fraction of the cost of training the original model \cite{yu2020cloudleak, yuan2020attack, tramer2016stealing}. 
\\
\newline
While these attacks can also target cloud based deployment, they become much easier to pull off in the case of edge deployment. Direct model stealing would require the attacker to gain access to the server infrastructure in the case of cloud deployment but in the case of edge deployment, the weights of the model are already downloaded to his device. Indirect model stealing is also possible with cloud deployments but they become much more feasible with edge deployment as now, the attacker does not need to launch thousands of network requests to query the model, instead he just evaluates the model locally. Where an attack like this might be detected in the case of cloud deployment because of the high volume of requests, it will likely remain undetected in the case of edge deployment.
\\
\newline
There are several protection mechanisms that might be considered to protect against model stealing attacks. First of all, encryption techniques can protect the model while it is downloaded or stored on the device. The model is then decrypted as it is loaded in memory, right before being used. Several frameworks such as the OpenVino toolkit \cite{gorbachev2019openvino} and Apple's CoreML \cite{marques2020machine} support encrypted models. This already provides a significant obstacle for direct model stealing. A disadvantage of this approach however is the increased computational cost caused by decrypting the model before use. The most secure approach is to use a Secure Processing Environment (SPE)  to handle model decryption or evaluation in an isolated protected subsystem on the device. This is however not always available on the low-end edge devices and if it is available, the isolation guarantees come at a price in terms of performance. A pragmatic solution is to evaluate only a part of the model on the trusted environment \cite{tramer2018slalom}.
\\
\newline
Another commonly discussed solution is watermarking. Here, additional information is stored in the weights of the model that can prove the ownership of the model. Watermarking does not prevent model theft, it can only be used after the fact as evidence. There are many different approaches to watermarking of deep neural network models \cite{nagai2018digital, zhang2018protecting, mellimi2021fast} . They are often compared in terms of the trade-off between fidelity, robustness and capacity \cite{li2021survey}. Here, fidelity refers to the impact on the model's performance on the original task. Robustness describes the resilience of the watermark against distortions such as those caused by pruning, finetuning  or deliberately overwriting the watermark. Finally, capacity indicates the size of the information that can be stored in the watermark itself. Most approaches embed the watermark during training by adding a term to the loss function that forces the model to embed additional information in the distribution of the model weights.  We distinguish between static and dynamic watermarking techniques that differ in the way the watermark can be read from the model \cite{li2021survey}. Static watermarking techniques embed the watermark into the weights of the model during training. This means that we need white-box access to the model to retrieve the watermark data itself. Dynamic watermarking techniques on the other hand embed the watermark implicitly by training the model to behave in a specific way for a carefully designed set of trigger inputs. This has the benefit that only black-box access to the output of the model is required. Watermarking techniques offer limited protection as they can only be used to prove that a model was stolen, they do not prevent theft. If the attacker uses the stolen weights privately, the theft might not even be discovered. For edge based deployment, the benefit of watermarking is that it does not incur a penalty in terms of computational cost or memory usage of the model. The disadvantage however is that the techniques usually already have to be applied during model training. TinyMLOps platforms  therefore have to keep track of the different versions of the model to associate different watermarks with different users.
\\
\newline
Watermarking offers no protection against indirect model stealing where the attacker repeatedly queries the model to generate training data pairs for his own model. There are two common families of solutions to protect against this: detecting stealing queries patterns and prediction poisoning. Detecting stealing queries patterns is a form of observability (see section \ref{sec:observability}). There are different techniques that analyze the distribution of sequential queries \cite{juuti2019prada} or that measure the information gain from different queries \cite{kesarwani2018model} to try to detect indirect model stealing. Prediction poisoning on the other hand takes a proactive approach by actively perturbing the outputs of the model that is returned to the user. These perturbations are carefully designed to retain the model accuracy while introducing sufficient noise to disturb the training process of a derivative model. Prediction poisoning can be as simple as rounding the confidence values \cite{tramer2016stealing}. More complicated approaches perturb the prediction in a way to guide the training process of the derivative model in the wrong direction \cite{orekondy2019prediction}. We would argue that the risk of indirect model stealing is higher for edge based applications than for cloud based applications and the model's owners might benefit greatly from protecting their intellectual property from these types of attacks. Although it is not supported yet by any of the TinyML frameworks, it seems feasible to perform stealing queries patterns detection and prediction poisoning on edge devices. 
\\
\newline
Other approaches to protect the intellectual property of machine learning models rely on homomorphic encryption \cite{gomez2018intellectual}, weight scrambling \cite{lin2020chaotic} or designing models that require a secret key to operate at their full potential \cite{chakraborty2020hardware}.
\\
\newline
All these types of protection mechanisms are tricky to implement and application developers might not have the time or background to implement them. They would therefore be a very valuable functionality of a TinyMLOps platform. 

\section{Verifiable execution on untrusted devices}
\label{sec:verify}
In large scale applications, the ML model is only a small part of the pipeline. The predictions of the model might be used to trigger other actions. We could for example use a face recognition model to authorize a payment. If the model is evaluated on cloud infrastructure controlled by the developers of the model, we can be sure that we are indeed working with the true predictions of the model. This becomes less trivial if the model is offloaded to the edge device. A malicious user might have changed the model or the predictions to trick the system into thinking that certain conditions are met. If this is a realistic concern for your application, it might be worthwhile to implement a \textit{verifiable computation} (VC) component. This allows an agent to provably (and cheaply) verify that an untrusted
party has performed the computations correctly \cite{ghodsi2017safetynets}. There are several solution to do this but the most interesting approaches evaluate the model and provide a small (in terms of number of bits) mathematical proof of the correctness of the result. The next stages in the pipeline (e.g. the component that authorizes the payment) can then use this to verify that this was indeed the result of an unmodified model. Note that this does not guarantee the accuracy of the prediction, it is still possible that the model has made a mistake or that the user has provided a forged input to the model. It merely guarantees that the prediction was indeed the result of the unmodified model. A major difficulty when implementing such a verification step in TinyML is the overhead caused by generating the proof. Sometimes this overhead can even be larger than the time needed to evaluate the original function \cite{walfish2015verifying}. Although recent innovations have reduced this overhead to about 5\% of the execution time of a model for MNIST digit recognition and TIMIT speech recognition tasks\cite{ghodsi2017safetynets}, it still remains extremely expensive for large scale models \cite{lee2020vcnn}.
\\
\newline 
Alternative solutions for verifiable execution require the support of Secure Processing Environments (SPE) such as Intel SGX or ARM TrustZone but these increase the cost of the edge device and can also reduce the performance \cite{tramer2018slalom}. An especially promising approach is this area is MLCapsule \cite{hanzlik2021mlcapsule} which provides a proof-of-concept on Intel SGX. Modern neural networks for image classification such as MobileNet have an overhead of around 2X when implemented using their approach.
\\
\newline
Similar to the model confidentiality techniques from previous section, these techniques are not trivial to implement securely. A TinyMLOps platform that automatically generates verifiable modules would therefore be of great value.

\section{Conclusion and future work}
\label{sec:conclude}
In this paper, we gave an overview of challenges that might arise when a developer opts for an edge based deployment of his ML application as compared to a cloud based deployment. We focused mainly on the operational side and listed several challenges that are either caused by the distributed edge deployment or that are complicated by it. As a field, TinyML is still very young with most of the tools and frameworks still in their early stages. With this document, we hope to inspire and guide the development of TinyMLOps platforms that will make TinyML accessible to developers and scalable to billions of edge devices.

\newpage
\bibliographystyle{IEEEtran}
\bibliography{references}

\begin{thebibliography}{10}
\providecommand{\url}[1]{#1}
\csname url@samestyle\endcsname
\providecommand{\newblock}{\relax}
\providecommand{\bibinfo}[2]{#2}
\providecommand{\BIBentrySTDinterwordspacing}{\spaceskip=0pt\relax}
\providecommand{\BIBentryALTinterwordstretchfactor}{4}
\providecommand{\BIBentryALTinterwordspacing}{\spaceskip=\fontdimen2\font plus
\BIBentryALTinterwordstretchfactor\fontdimen3\font minus
  \fontdimen4\font\relax}
\providecommand{\BIBforeignlanguage}[2]{{%
\expandafter\ifx\csname l@#1\endcsname\relax
\typeout{** WARNING: IEEEtran.bst: No hyphenation pattern has been}%
\typeout{** loaded for the language `#1'. Using the pattern for}%
\typeout{** the default language instead.}%
\else
\language=\csname l@#1\endcsname
\fi
#2}}
\providecommand{\BIBdecl}{\relax}
\BIBdecl

\bibitem{warden2019tinyml}
P.~Warden and D.~Situnayake, \emph{TinyML}.\hskip 1em plus 0.5em minus
  0.4em\relax O'Reilly Media, Incorporated, 2019.

\bibitem{dutta2021tinyml}
L.~Dutta and S.~Bharali, ``Tinyml meets iot: A comprehensive survey,''
  \emph{Internet of Things}, vol.~16, p. 100461, 2021.

\bibitem{han2015learning}
S.~Han, J.~Pool, J.~Tran, and W.~J. Dally, ``Learning both weights and
  connections for efficient neural networks,'' \emph{arXiv preprint
  arXiv:1506.02626}, 2015.

\bibitem{nagel2021white}
M.~Nagel, M.~Fournarakis, R.~A. Amjad, Y.~Bondarenko, M.~van Baalen, and
  T.~Blankevoort, ``A white paper on neural network quantization,'' \emph{arXiv
  preprint arXiv:2106.08295}, 2021.

\bibitem{yang2019snapshot}
C.~Yang, L.~Xie, C.~Su, and A.~L. Yuille, ``Snapshot distillation:
  Teacher-student optimization in one generation,'' in \emph{Proceedings of the
  IEEE/CVF Conference on Computer Vision and Pattern Recognition}, 2019, pp.
  2859--2868.

\bibitem{leroux2018iamnn}
S.~Leroux, P.~Molchanov, P.~Simoens, B.~Dhoedt, T.~Breuel, and J.~Kautz,
  ``Iamnn: Iterative and adaptive mobile neural network for efficient image
  classification,'' \emph{arXiv preprint arXiv:1804.10123}, 2018.

\bibitem{zoph2016neural}
B.~Zoph and Q.~V. Le, ``Neural architecture search with reinforcement
  learning,'' \emph{arXiv preprint arXiv:1611.01578}, 2016.

\bibitem{sze2017efficient}
V.~Sze, Y.-H. Chen, T.-J. Yang, and J.~S. Emer, ``Efficient processing of deep
  neural networks: A tutorial and survey,'' \emph{Proceedings of the IEEE},
  vol. 105, no.~12, pp. 2295--2329, 2017.

\bibitem{moons201714}
B.~Moons, R.~Uytterhoeven, W.~Dehaene, and M.~Verhelst, ``14.5 envision: A
  0.26-to-10tops/w subword-parallel dynamic-voltage-accuracy-frequency-scalable
  convolutional neural network processor in 28nm fdsoi,'' in \emph{2017 IEEE
  International Solid-State Circuits Conference (ISSCC)}.\hskip 1em plus 0.5em
  minus 0.4em\relax IEEE, 2017, pp. 246--247.

\bibitem{xu2020edge}
D.~Xu, T.~Li, Y.~Li, X.~Su, S.~Tarkoma, T.~Jiang, J.~Crowcroft, and P.~Hui,
  ``Edge intelligence: Architectures, challenges, and applications,''
  \emph{arXiv preprint arXiv:2003.12172}, 2020.

\bibitem{verhelst2017embedded}
M.~Verhelst and B.~Moons, ``Embedded deep neural network processing:
  Algorithmic and processor techniques bring deep learning to iot and edge
  devices,'' \emph{IEEE Solid-State Circuits Magazine}, vol.~9, no.~4, pp.
  55--65, 2017.

\bibitem{gartner}
``Gartner study on mlops,''
  \url{https://www.gartner.com/en/newsroom/press-releases/2018-02-13-gartner-says-nearly-half-of-cios-are-planning-to-deploy-artificial-intelligence
  }, accessed: 2021-12-11.

\bibitem{sculley2015hidden}
D.~Sculley, G.~Holt, D.~Golovin, E.~Davydov, T.~Phillips, D.~Ebner,
  V.~Chaudhary, M.~Young, J.-F. Crespo, and D.~Dennison, ``Hidden technical
  debt in machine learning systems,'' \emph{Advances in neural information
  processing systems}, vol.~28, pp. 2503--2511, 2015.

\bibitem{suryanarayana2014refactoring}
G.~Suryanarayana, G.~Samarthyam, and T.~Sharma, \emph{Refactoring for software
  design smells: managing technical debt}.\hskip 1em plus 0.5em minus
  0.4em\relax Morgan Kaufmann, 2014.

\bibitem{muralidhar2021using}
N.~Muralidhar, S.~Muthiah, P.~Butler, M.~Jain, Y.~Yu, K.~Burne, W.~Li,
  D.~Jones, P.~Arunachalam, H.~S. McCormick \emph{et~al.}, ``Using antipatterns
  to avoid mlops mistakes,'' \emph{arXiv preprint arXiv:2107.00079}, 2021.

\bibitem{deloite}
``Deloite study on mlops,''
  \url{https://www2.deloitte.com/content/dam/\\insights/articles/6730\_TT-Landing-page/DI\_2021-Tech-Trends.pdf},
  accessed: 2021-12-11.

\bibitem{leroux2019multi}
S.~Leroux, S.~Bohez, E.~De~Coninck, P.~Van~Molle, B.~Vankeirsbilck,
  T.~Verbelen, P.~Simoens, and B.~Dhoedt, ``Multi-fidelity deep neural networks
  for adaptive inference in the internet of multimedia things,'' \emph{Future
  Generation Computer Systems}, vol.~97, pp. 355--360, 2019.

\bibitem{vanhoucke2011improving}
V.~Vanhoucke, A.~Senior, and M.~Z. Mao, ``Improving the speed of neural
  networks on cpus,'' 2011.

\bibitem{nayak2019bit}
P.~Nayak, D.~Zhang, and S.~Chai, ``Bit efficient quantization for deep neural
  networks,'' \emph{arXiv preprint arXiv:1910.04877}, 2019.

\bibitem{venkatesh2017accelerating}
G.~Venkatesh, E.~Nurvitadhi, and D.~Marr, ``Accelerating deep convolutional
  networks using low-precision and sparsity,'' in \emph{2017 IEEE International
  Conference on Acoustics, Speech and Signal Processing (ICASSP)}.\hskip 1em
  plus 0.5em minus 0.4em\relax IEEE, 2017, pp. 2861--2865.

\bibitem{courbariaux2016binarized}
M.~Courbariaux, I.~Hubara, D.~Soudry, R.~El-Yaniv, and Y.~Bengio, ``Binarized
  neural networks: Training deep neural networks with weights and activations
  constrained to+ 1 or-1,'' \emph{arXiv preprint arXiv:1602.02830}, 2016.

\bibitem{judd2016stripes}
P.~Judd, J.~Albericio, T.~Hetherington, T.~M. Aamodt, and A.~Moshovos,
  ``Stripes: Bit-serial deep neural network computing,'' in \emph{2016 49th
  Annual IEEE/ACM International Symposium on Microarchitecture (MICRO)}.\hskip
  1em plus 0.5em minus 0.4em\relax IEEE, 2016, pp. 1--12.

\bibitem{haas2017bringing}
A.~Haas, A.~Rossberg, D.~L. Schuff, B.~L. Titzer, M.~Holman, D.~Gohman,
  L.~Wagner, A.~Zakai, and J.~Bastien, ``Bringing the web up to speed with
  webassembly,'' in \emph{Proceedings of the 38th ACM SIGPLAN Conference on
  Programming Language Design and Implementation}, 2017, pp. 185--200.

\bibitem{lootus2020vm}
M.~Lootus, K.~Thakore, S.~Leroux, G.~Trooskens, A.~Sharma, and H.~Ly, ``A
  vm/containerized approach for scaling tinyml applications,'' in
  \emph{Research Symposium on Tiny Machine Learning}, 2020.

\bibitem{shankar2021towards}
S.~Shankar and A.~Parameswaran, ``Towards observability for machine learning
  pipelines,'' \emph{arXiv preprint arXiv:2108.13557}, 2021.

\bibitem{arize}
``Arize ml observability solution,''
  \url{https://arize.com/platform-overview/}, accessed: 2021-12-11.

\bibitem{whylabs}
``Whylabs ml observability solution,'' \url{https://whylabs.ai/}, accessed:
  2021-12-11.

\bibitem{evidently}
``evidently ai monitoring,'' \url{https://evidentlyai.com/}, accessed:
  2021-12-11.

\bibitem{cloudvision}
``Google's cloud vision rates,'' \url{https://cloud.google.com/vision/pricing},
  accessed: 2021-12-11.

\bibitem{christodorescu2020towards}
M.~Christodorescu, W.~C. Gu, R.~Kumaresan, M.~Minaei, M.~Ozdayi, B.~Price,
  S.~Raghuraman, M.~Saad, C.~Sheffield, M.~Xu \emph{et~al.}, ``Towards a
  two-tier hierarchical infrastructure: An offline payment system for central
  bank digital currencies,'' \emph{arXiv preprint arXiv:2012.08003}, 2020.

\bibitem{thai2021does}
A.~Thai, S.~Stojanov, I.~Rehg, and J.~M. Rehg, ``Does continual learning=
  catastrophic forgetting?'' \emph{arXiv preprint arXiv:2101.07295}, 2021.

\bibitem{mcmahan2017communication}
B.~McMahan, E.~Moore, D.~Ramage, S.~Hampson, and B.~A. y~Arcas,
  ``Communication-efficient learning of deep networks from decentralized
  data,'' in \emph{Artificial intelligence and statistics}.\hskip 1em plus
  0.5em minus 0.4em\relax PMLR, 2017, pp. 1273--1282.

\bibitem{zhu2021federated}
H.~Zhu, J.~Xu, S.~Liu, and Y.~Jin, ``Federated learning on non-iid data: A
  survey,'' \emph{arXiv preprint arXiv:2106.06843}, 2021.

\bibitem{bonawitz2019towards}
K.~Bonawitz, H.~Eichner, W.~Grieskamp, D.~Huba, A.~Ingerman, V.~Ivanov,
  C.~Kiddon, J.~Kone{\v{c}}n{\`y}, S.~Mazzocchi, H.~B. McMahan \emph{et~al.},
  ``Towards federated learning at scale: System design,'' \emph{arXiv preprint
  arXiv:1902.01046}, 2019.

\bibitem{luo2021cost}
B.~Luo, X.~Li, S.~Wang, J.~Huang, and L.~Tassiulas, ``Cost-effective federated
  learning design,'' in \emph{IEEE INFOCOM 2021-IEEE Conference on Computer
  Communications}.\hskip 1em plus 0.5em minus 0.4em\relax IEEE, 2021, pp.
  1--10.

\bibitem{yao2019federated}
X.~Yao, T.~Huang, C.~Wu, R.-X. Zhang, and L.~Sun, ``Federated learning with
  additional mechanisms on clients to reduce communication costs,'' \emph{arXiv
  preprint arXiv:1908.05891}, 2019.

\bibitem{chen2019communication}
Y.~Chen, X.~Sun, and Y.~Jin, ``Communication-efficient federated deep learning
  with layerwise asynchronous model update and temporally weighted
  aggregation,'' \emph{IEEE transactions on neural networks and learning
  systems}, vol.~31, no.~10, pp. 4229--4238, 2019.

\bibitem{yao2018two}
X.~Yao, C.~Huang, and L.~Sun, ``Two-stream federated learning: Reduce the
  communication costs,'' in \emph{2018 IEEE Visual Communications and Image
  Processing (VCIP)}.\hskip 1em plus 0.5em minus 0.4em\relax IEEE, 2018, pp.
  1--4.

\bibitem{mills2019communication}
J.~Mills, J.~Hu, and G.~Min, ``Communication-efficient federated learning for
  wireless edge intelligence in iot,'' \emph{IEEE Internet of Things Journal},
  vol.~7, no.~7, pp. 5986--5994, 2019.

\bibitem{xu2020ternary}
J.~Xu, W.~Du, Y.~Jin, W.~He, and R.~Cheng, ``Ternary compression for
  communication-efficient federated learning,'' \emph{IEEE Transactions on
  Neural Networks and Learning Systems}, 2020.

\bibitem{duan2019astraea}
M.~Duan, D.~Liu, X.~Chen, Y.~Tan, J.~Ren, L.~Qiao, and L.~Liang, ``Astraea:
  Self-balancing federated learning for improving classification accuracy of
  mobile deep learning applications,'' in \emph{2019 IEEE 37th international
  conference on computer design (ICCD)}.\hskip 1em plus 0.5em minus 0.4em\relax
  IEEE, 2019, pp. 246--254.

\bibitem{lim2020federated}
W.~Y.~B. Lim, N.~C. Luong, D.~T. Hoang, Y.~Jiao, Y.-C. Liang, Q.~Yang,
  D.~Niyato, and C.~Miao, ``Federated learning in mobile edge networks: A
  comprehensive survey,'' \emph{IEEE Communications Surveys \& Tutorials},
  vol.~22, no.~3, pp. 2031--2063, 2020.

\bibitem{semwal2020fedperf}
T.~Semwal, A.~Mulay, and A.~M. Agrawal, ``Fedperf: A practitioners’ guide to
  performance of federated learning algorithms,'' 2020.

\bibitem{diao2021semifl}
E.~Diao, J.~Ding, and V.~Tarokh, ``Semifl: Communication efficient
  semi-supervised federated learning with unlabeled clients,'' \emph{arXiv
  preprint arXiv:2106.01432}, 2021.

\bibitem{jin2020towards}
Y.~Jin, X.~Wei, Y.~Liu, and Q.~Yang, ``Towards utilizing unlabeled data in
  federated learning: A survey and prospective,'' \emph{arXiv preprint
  arXiv:2002.11545}, 2020.

\bibitem{jeong2020federated}
W.~Jeong, J.~Yoon, E.~Yang, and S.~J. Hwang, ``Federated semi-supervised
  learning with inter-client consistency,'' \emph{arXiv e-prints}, pp.
  arXiv--2006, 2020.

\bibitem{albaseer2020exploiting}
A.~Albaseer, B.~S. Ciftler, M.~Abdallah, and A.~Al-Fuqaha, ``Exploiting
  unlabeled data in smart cities using federated learning,'' \emph{arXiv
  preprint arXiv:2001.04030}, 2020.

\bibitem{itahara2020distillation}
S.~Itahara, T.~Nishio, Y.~Koda, M.~Morikura, and K.~Yamamoto,
  ``Distillation-based semi-supervised federated learning for
  communication-efficient collaborative training with non-iid private data,''
  \emph{arXiv preprint arXiv:2008.06180}, 2020.

\bibitem{zhang2020federated}
F.~Zhang, K.~Kuang, Z.~You, T.~Shen, J.~Xiao, Y.~Zhang, C.~Wu, Y.~Zhuang, and
  X.~Li, ``Federated unsupervised representation learning,'' \emph{arXiv
  preprint arXiv:2010.08982}, 2020.

\bibitem{tzinis2021separate}
E.~Tzinis, J.~Casebeer, Z.~Wang, and P.~Smaragdis, ``Separate but together:
  Unsupervised federated learning for speech enhancement from non-iid data,''
  \emph{arXiv preprint arXiv:2105.04727}, 2021.

\bibitem{kannadasan2019personalized}
M.~R. Kannadasan and G.~Aslanyan, ``Personalized query auto-completion through
  a lightweight representation of the user context,'' \emph{arXiv preprint
  arXiv:1905.01386}, 2019.

\bibitem{anaissi2021personalized}
A.~Anaissi and B.~Suleiman, ``A personalized federated learning algorithm: an
  application in anomaly detection,'' \emph{arXiv preprint arXiv:2111.02627},
  2021.

\bibitem{LEROUX2022107763}
\BIBentryALTinterwordspacing
S.~Leroux, B.~Li, and P.~Simoens, ``Automated training of location-specific
  edge models for traffic counting,'' \emph{Computers and Electrical
  Engineering}, vol.~99, p. 107763, 2022. [Online]. Available:
  \url{https://www.sciencedirect.com/science/article/pii/S0045790622000672}
\BIBentrySTDinterwordspacing

\bibitem{li2020survey}
W.~Li and M.~Liewig, ``A survey of ai accelerators for edge environment,'' in
  \emph{World Conference on Information Systems and Technologies}.\hskip 1em
  plus 0.5em minus 0.4em\relax Springer, 2020, pp. 35--44.

\bibitem{onnx}
``Onnx: Open neural network exchange,'' \url{https://onnx.ai/}, accessed:
  2021-12-11.

\bibitem{nnef}
``Nnef: Neural network exchange format,'' \url{https://www.khronos.org/nnef},
  accessed: 2021-12-11.

\bibitem{chen2018tvm}
T.~Chen, T.~Moreau, Z.~Jiang, L.~Zheng, E.~Yan, H.~Shen, M.~Cowan, L.~Wang,
  Y.~Hu, L.~Ceze \emph{et~al.}, ``$\{$TVM$\}$: An automated end-to-end
  optimizing compiler for deep learning,'' in \emph{13th $\{$USENIX$\}$
  Symposium on Operating Systems Design and Implementation ($\{$OSDI$\}$ 18)},
  2018, pp. 578--594.

\bibitem{gorbachev2019openvino}
Y.~Gorbachev, M.~Fedorov, I.~Slavutin, A.~Tugarev, M.~Fatekhov, and Y.~Tarkan,
  ``Openvino deep learning workbench: Comprehensive analysis and tuning of
  neural networks inference,'' in \emph{Proceedings of the IEEE/CVF
  International Conference on Computer Vision Workshops}, 2019, pp. 0--0.

\bibitem{kenning}
``Kenning edge ai framework,''
  \url{https://antmicro.com/blog/2021/06/kenning-edge-ai-framework/ },
  accessed: 2021-12-11.

\bibitem{de2018dianne}
E.~De~Coninck, S.~Bohez, S.~Leroux, T.~Verbelen, B.~Vankeirsbilck, P.~Simoens,
  and B.~Dhoedt, ``Dianne: a modular framework for designing, training and
  deploying deep neural networks on heterogeneous distributed infrastructure,''
  \emph{Journal of Systems and Software}, vol. 141, pp. 52--65, 2018.

\bibitem{zhu2020toward}
G.~Zhu, D.~Liu, Y.~Du, C.~You, J.~Zhang, and K.~Huang, ``Toward an intelligent
  edge: Wireless communication meets machine learning,'' \emph{IEEE
  communications magazine}, vol.~58, no.~1, pp. 19--25, 2020.

\bibitem{leroux2017cascading}
S.~Leroux, S.~Bohez, E.~De~Coninck, T.~Verbelen, B.~Vankeirsbilck, P.~Simoens,
  and B.~Dhoedt, ``The cascading neural network: building the internet of smart
  things,'' \emph{Knowledge and Information Systems}, vol.~52, no.~3, pp.
  791--814, 2017.

\bibitem{banitalebi2021auto}
A.~Banitalebi-Dehkordi, N.~Vedula, J.~Pei, F.~Xia, L.~Wang, and Y.~Zhang,
  ``Auto-split: A general framework of collaborative edge-cloud ai,'' in
  \emph{Proceedings of the 27th ACM SIGKDD Conference on Knowledge Discovery \&
  Data Mining}, 2021, pp. 2543--2553.

\bibitem{grulich2018collaborative}
P.~M. Grulich and F.~Nawab, ``Collaborative edge and cloud neural networks for
  real-time video processing,'' \emph{Proceedings of the VLDB Endowment},
  vol.~11, no.~12, pp. 2046--2049, 2018.

\bibitem{tuli2021splitplace}
S.~Tuli, ``Splitplace: Intelligent placement of split neural nets in mobile
  edge environments,'' \emph{arXiv preprint arXiv:2110.04841}, 2021.

\bibitem{gpt3}
``Demystifying gpt-3,'' \url{https://lambdalabs.com/blog/demystifying-gpt-3/},
  accessed: 2021-12-11.

\bibitem{shokri2017membership}
R.~Shokri, M.~Stronati, C.~Song, and V.~Shmatikov, ``Membership inference
  attacks against machine learning models,'' in \emph{2017 IEEE Symposium on
  Security and Privacy (SP)}.\hskip 1em plus 0.5em minus 0.4em\relax IEEE,
  2017, pp. 3--18.

\bibitem{fredrikson2015model}
M.~Fredrikson, S.~Jha, and T.~Ristenpart, ``Model inversion attacks that
  exploit confidence information and basic countermeasures,'' in
  \emph{Proceedings of the 22nd ACM SIGSAC conference on computer and
  communications security}, 2015, pp. 1322--1333.

\bibitem{yu2020cloudleak}
H.~Yu, K.~Yang, T.~Zhang, Y.-Y. Tsai, T.-Y. Ho, and Y.~Jin, ``Cloudleak:
  Large-scale deep learning models stealing through adversarial examples.'' in
  \emph{NDSS}, 2020.

\bibitem{yuan2020attack}
X.~Yuan, L.~Ding, L.~Zhang, X.~Li, and D.~Wu, ``Es attack: Model stealing
  against deep neural networks without data hurdles,'' \emph{arXiv preprint
  arXiv:2009.09560}, 2020.

\bibitem{tramer2016stealing}
F.~Tram{\`e}r, F.~Zhang, A.~Juels, M.~K. Reiter, and T.~Ristenpart, ``Stealing
  machine learning models via prediction apis,'' in \emph{25th $\{$USENIX$\}$
  Security Symposium ($\{$USENIX$\}$ Security 16)}, 2016, pp. 601--618.

\bibitem{marques2020machine}
O.~Marques, ``Machine learning with core ml,'' in \emph{Image Processing and
  Computer Vision in iOS}.\hskip 1em plus 0.5em minus 0.4em\relax Springer,
  2020, pp. 29--40.

\bibitem{tramer2018slalom}
F.~Tramer and D.~Boneh, ``Slalom: Fast, verifiable and private execution of
  neural networks in trusted hardware,'' \emph{arXiv preprint
  arXiv:1806.03287}, 2018.

\bibitem{nagai2018digital}
Y.~Nagai, Y.~Uchida, S.~Sakazawa, and S.~Satoh, ``Digital watermarking for deep
  neural networks,'' \emph{International Journal of Multimedia Information
  Retrieval}, vol.~7, no.~1, pp. 3--16, 2018.

\bibitem{zhang2018protecting}
J.~Zhang, Z.~Gu, J.~Jang, H.~Wu, M.~P. Stoecklin, H.~Huang, and I.~Molloy,
  ``Protecting intellectual property of deep neural networks with
  watermarking,'' in \emph{Proceedings of the 2018 on Asia Conference on
  Computer and Communications Security}, 2018, pp. 159--172.

\bibitem{mellimi2021fast}
S.~Mellimi, V.~Rajput, I.~A. Ansari, and C.~W. Ahn, ``A fast and efficient
  image watermarking scheme based on deep neural network,'' \emph{Pattern
  Recognition Letters}, vol. 151, pp. 222--228, 2021.

\bibitem{li2021survey}
Y.~Li, H.~Wang, and M.~Barni, ``A survey of deep neural network watermarking
  techniques,'' \emph{arXiv preprint arXiv:2103.09274}, 2021.

\bibitem{juuti2019prada}
M.~Juuti, S.~Szyller, S.~Marchal, and N.~Asokan, ``Prada: protecting against
  dnn model stealing attacks,'' in \emph{2019 IEEE European Symposium on
  Security and Privacy (EuroS\&P)}.\hskip 1em plus 0.5em minus 0.4em\relax
  IEEE, 2019, pp. 512--527.

\bibitem{kesarwani2018model}
M.~Kesarwani, B.~Mukhoty, V.~Arya, and S.~Mehta, ``Model extraction warning in
  mlaas paradigm,'' in \emph{Proceedings of the 34th Annual Computer Security
  Applications Conference}, 2018, pp. 371--380.

\bibitem{orekondy2019prediction}
T.~Orekondy, B.~Schiele, and M.~Fritz, ``Prediction poisoning: Towards defenses
  against dnn model stealing attacks,'' \emph{arXiv preprint arXiv:1906.10908},
  2019.

\bibitem{gomez2018intellectual}
L.~Gomez, A.~Ibarrondo, J.~M{\'a}rquez, and P.~Duverger, ``Intellectual
  property protection for distributed neural networks,'' 2018.

\bibitem{lin2020chaotic}
N.~Lin, X.~Chen, H.~Lu, and X.~Li, ``Chaotic weights: A novel approach to
  protect intellectual property of deep neural networks,'' \emph{IEEE
  Transactions on Computer-Aided Design of Integrated Circuits and Systems},
  vol.~40, no.~7, pp. 1327--1339, 2020.

\bibitem{chakraborty2020hardware}
A.~Chakraborty, A.~Mondai, and A.~Srivastava, ``Hardware-assisted intellectual
  property protection of deep learning models,'' in \emph{2020 57th ACM/IEEE
  Design Automation Conference (DAC)}.\hskip 1em plus 0.5em minus 0.4em\relax
  IEEE, 2020, pp. 1--6.

\bibitem{ghodsi2017safetynets}
Z.~Ghodsi, T.~Gu, and S.~Garg, ``Safetynets: Verifiable execution of deep
  neural networks on an untrusted cloud,'' \emph{arXiv preprint
  arXiv:1706.10268}, 2017.

\bibitem{walfish2015verifying}
M.~Walfish and A.~J. Blumberg, ``Verifying computations without reexecuting
  them,'' \emph{Communications of the ACM}, vol.~58, no.~2, pp. 74--84, 2015.

\bibitem{lee2020vcnn}
S.~Lee, H.~Ko, J.~Kim, and H.~Oh, ``vcnn: Verifiable convolutional neural
  network based on zk-snarks,'' Cryptology ePrint Archive, Report 2020/584.
  https://eprint. iacr. org/2020/584, Tech. Rep., 2020.

\bibitem{hanzlik2021mlcapsule}
L.~Hanzlik, Y.~Zhang, K.~Grosse, A.~Salem, M.~Augustin, M.~Backes, and
  M.~Fritz, ``Mlcapsule: Guarded offline deployment of machine learning as a
  service,'' in \emph{Proceedings of the IEEE/CVF Conference on Computer Vision
  and Pattern Recognition}, 2021, pp. 3300--3309.

\end{thebibliography}

\end{document}